\documentclass[sigconf]{acmart}
\usepackage{subfig}
\usepackage{multirow}
\usepackage[normalem]{ulem}
\useunder{\uline}{\ul}{}
\usepackage{arydshln} 
\usepackage{booktabs}
\usepackage{bm}
\usepackage{hyphenat}

\AtBeginDocument{%
  \providecommand\BibTeX{{%
    \normalfont B\kern-0.5em{\scshape i\kern-0.25em b}\kern-0.8em\TeX}}}

\copyrightyear{2026}
\acmYear{2026}
\setcopyright{acmlicensed}\acmConference[MM '26]{Proceedings of the 34nd ACM International Conference on Multimedia}{November 10-14, 2026}{Rio de Janeiro, Brazil}
\acmBooktitle{Proceedings of the 34nd ACM International Conference on Multimedia (MM '26), November 10-14, 2026, Rio de Janeiro, Brazil}
\acmDOI{xxxx}
\acmISBN{xxxx}

\begin{document}

\title{Listening Deepfake Detection: A New Perspective Beyond Speaking-Centric Forgery Analysis}

\author{Miao Liu}
\orcid{0000-0002-2039-2051}
\affiliation{%
	\institution{Beijing Institute of Technology}
	\city{Beijing}
	\country{China}
}
\email{3120225410@bit.edu.cn}

\author{Fangda Wei}
\orcid{0009-0008-6248-0018}
\affiliation{%
	\institution{Beijing Institute of Technology}
	\city{Beijing}
	\country{China}
}
\email{3220240872@bit.edu.cn}

\author{Jing Wang}
\orcid{0000-0002-3653-9951}
\authornote{Corresponding author.}
\affiliation{%
	\institution{Beijing Institute of Technology}
	\city{Beijing}
	\country{China}
}
\email{wangjing@bit.edu.cn}

\author{Xinyuan Qian}
\orcid{0000-0002-9511-6713}
\affiliation{%
	\institution{University of Science and Technology Beijing}
	\city{Beijing}
	\country{China}
}
\email{qianxy@ustb.edu.cn}


\renewcommand{\shortauthors}{Liu et al.}

\begin{abstract}
Existing deepfake detection research has primarily focused on scenarios where the manipulated subject is actively speaking, i.e., generating fabricated content by altering the speaker’s appearance or voice. However, in realistic interaction settings, attackers often alternate between falsifying speaking and listening states to mislead their targets, thereby enhancing the realism and persuasiveness of the scenario. Although the detection of 'listening deepfakes' remains largely unexplored and is hindered by a scarcity of both datasets and methodologies, the relatively limited quality of synthesized listening reactions presents an excellent breakthrough opportunity for current deepfake detection efforts. 
In this paper, we present the task of Listening Deepfake Detection (LDD). We introduce ListenForge, the first dataset specifically designed for this task, constructed using five Listening Head Generation (LHG) methods. To address the distinctive characteristics of listening forgeries, we propose MANet, a Motion-aware and Audio-guided Network that captures subtle motion inconsistencies in listener videos while leveraging speaker's audio semantics to guided cross-modal fusion. Extensive experiments demonstrate that existing Speaking Deepfake Detection (SDD) models perform poorly in listening scenarios. In contrast, MANet achieves significantly superior performance on ListenForge. Our work highlights the necessity of rethinking deepfake detection beyond the traditional speaking-centric paradigm and opens new directions for multimodal forgery analysis in interactive communication settings.  The dataset and code are available at  \href{https://anonymous.4open.science/r/LDD-B4CB}{https://anonymous.4open.science/r/LDD-B4CB}.

\end{abstract}



\maketitle

\begin{figure}
	\centering
	\includegraphics[width=8cm]{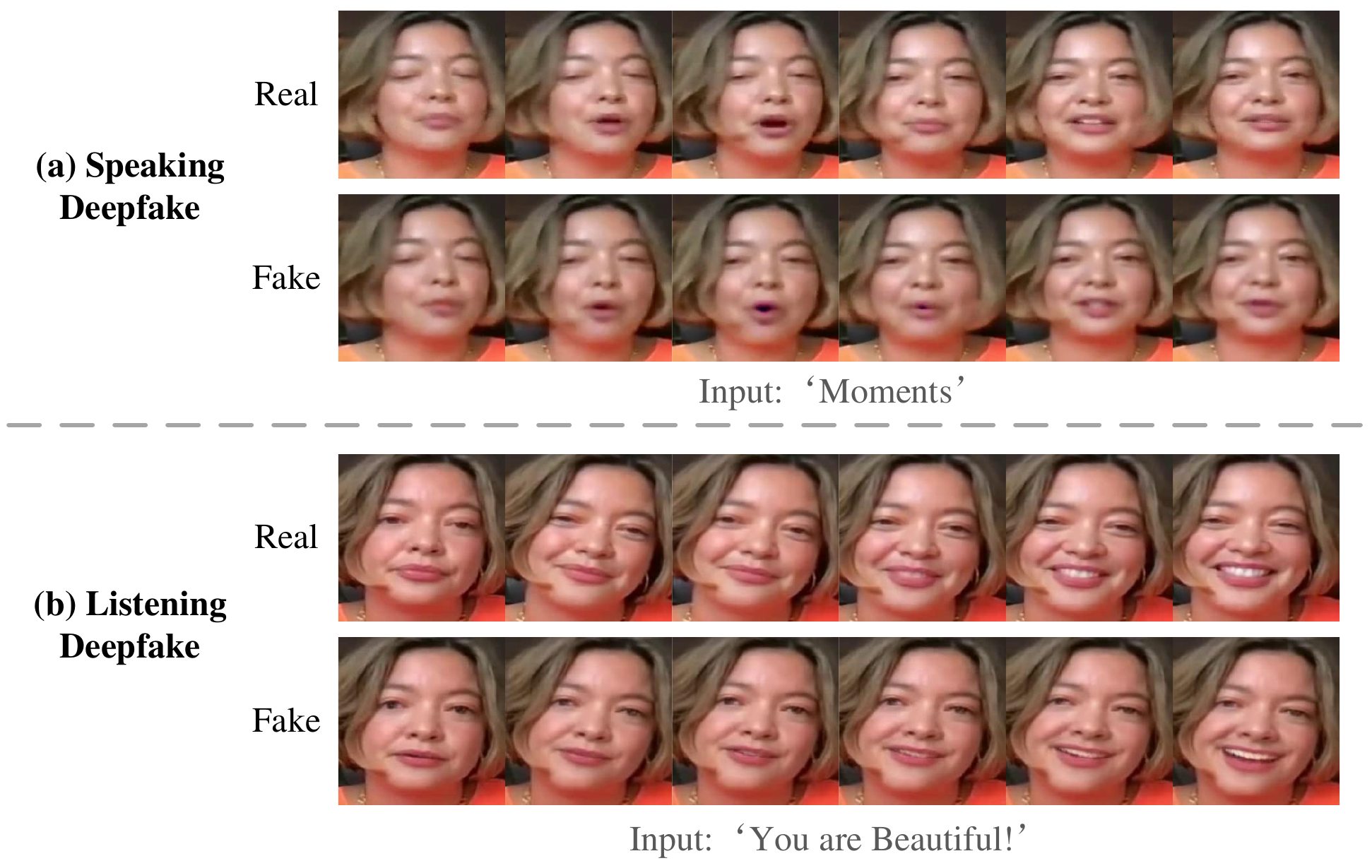}
	\caption{A visualization comparison between common speaking deepfakes and our studied listening
		deepfakes. The former relies on the synchronization between the lip region and the given audio during speaking, while the latter manifests forged facial movements such as smiling, nodding, or shaking the head during listening.}
	\label{f1}
\end{figure}

\section{Introduction}

Deepfake technology, driven by rapid advances in Artificial Intelligence Generated Content (AIGC) \cite{Review,cai2022devit,xu2026emotionally,yang2023diffusion}, has become increasingly widespread in recent years. This fast-evolving technology holds the potential to transform content creation by enabling the generation of customizable and scalable materials across diverse domains. Meanwhile, deepfake detection methods have emerged to mitigate the misuse of deepfake technology.

Existing deepfake detection research mainly focuses on scenarios where the manipulated subject is  actively speaking. As shown in Figure \ref{f1}a, this type of forgery primarily targets the synchronization between facial movements and the driving speech. Specifically, it manipulates the speaker's lip movements, expressions, and speech content to ensure audiovisual consistency in the generated video. However, in real-world interactive scenarios, such as online video conferences or social media video chats, the roles of the participants dynamically alternate between speaker and listener. To deceive their counterparts, criminals must often spoof not only the speaking state but also generate real-time facial feedback as a listener based on the victim's spoken content. Such highly synchronized interactive forgeries significantly amplify the realism and persuasiveness of the deceptive scenarios.


As shown in Figure \ref{f1}b, different from forgery methods targeting the speaking state, "listening deepfake" focuses on the manipulated subject's listening state. It leverages Listening Head Generation (LHG) technology to synthesize listener behaviors—such as facial expression variations and head movements—that correspond to the speaker's audio content during an interaction. By targeting the listener for synthesis, this technology constructively enhances virtual interactive experiences and demonstrates broad application potential across entertainment sectors, including digital humans, film and television production, gaming, and content creation. Simultaneously, however, this form of interactive deepfake substantially amplifies the immersion and credibility of deceptive scenarios, thereby significantly elevating the risk of its exploitation for malicious purposes, such as identity fraud and personalized disinformation campaigns.
Nevertheless, despite the vital role listeners play in genuine interactions, this aspect has long remained under-explored in deepfake detection research. Consequently, developing an efficient forgery detection mechanism tailored to the manipulated subject's listening state has emerged as a critical and urgent challenge.


Compared to mature and highly realistic speaker forgery techniques, current listener forgery technology remains in its infancy. The LHG method only emerged around 2022 \cite{zhou2022responsive}. Despite rapid advancements driven by various GANs and diffusion models, its overall research maturity still lags significantly behind that of Speaking Head Generation (SHG). Furthermore, synthesizing a listener presents greater challenges. In terms of facial micro-expressions, head poses, and attentional biases, the synthesized listeners often struggle to align deeply with the semantic context \cite{l2l}, making them highly susceptible to exposing anomalous cues. Consequently, it is reasonable to infer that forgeries in the listening state are more identifiable, potentially serving as a weak link within the current deepfake scenario.
Moreover, listening forgery attacks exhibit unique and highly discriminative characteristics in their manifestations, rendering them difficult to detect using traditional Speaking Deepfake Detection (SDD) methods. On one hand, existing SDD methods typically focus on exploiting inconsistency cues during speech, such as audio-visual inconsistency between lip movements and audio \cite{yang2020preventing}. However, such cues are entirely absent in listener-centric synthesis scenarios. On the other hand, the temporal alignment in this context shifts from the strong constraint of strict lip-syncing to a weaker requirement of semantic consistency. For instance, when a speaker tells a joke, the synthesized listener is expected to exhibit an appropriate smile or reaction.
These distinct attributes further underscore the imperative need to develop deepfake detection methods specifically designed for the manipulated subject's listening state, namely, Listening Deepfake Detection (LDD).


Meanwhile, the LDD task faces a severe scarcity of datasets. Although the SDD domain has amassed abundant forgery datasets, as previously discussed, these resources cannot be directly applied to the LDD task. Since existing research lacks attention to the listening deepfakes, and has not carried out corresponding dataset construction work, developing a dedicated dataset for this emerging scenario is crucial for the development of LDD. To this end, we introduces ListenForge, a listening forgery dataset specifically designed for training and evaluating LDD models. To the best of our knowledge, this is the first dataset explicitly constructed for the LDD scenario.


In this paper, for the first time, we propose a Motion-aware and Audio-guided Network (MANet) for the LDD task. In MANet, the motion-aware module captures subtle temporal inconsistencies and facial expression cues from listener videos, which are critical to assessing the authenticity of listener content. The audio-guided module extracts the correlation between speaker's audio semantics and the listener's visual responses, enabling the system to capture the contextual interaction between the speaker and the target listener. Together, these modules constitute a unified framework highly effective at detecting listening deepfakes.
The contributions of our work can be summarized as follows:
\begin{itemize}
	\item We first introduce the listening deepfake detection task, a problem that has been largely overlooked in existing research. Furthermore, we construct ListenForge, the first listening forgery dataset, to facilitate systematic investigation of this task.
	\item We propose MANet, which utilizes motion information and exploits cross-modal inconsistency for listening deepfake detection.
	\item We conduct extensive experiments on it alongside other forgery datasets to demonstrate that our method outperforms the existing deepfake detection methods in the LDD task.
\end{itemize}



\begin{figure*}[t]
	\centering
	\includegraphics[width=17cm]{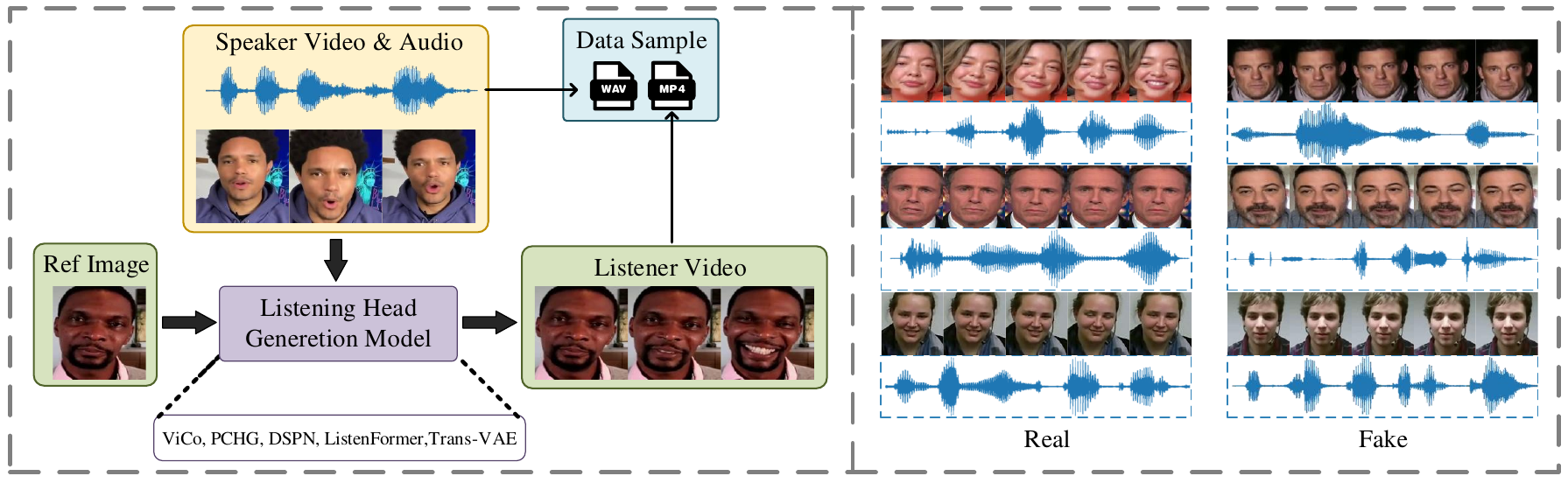}
	 \vspace{-0.2cm}
	\caption{ListenForge dataset construction. Utilizing five methods, we generated realistic listening videos accompanied by the speaker’s audio. The figure on the right illustrates a subset of real and fake samples from ListenForge dataset.}
	\label{f2}
\end{figure*}

\section{Related Work}

\subsection{Speaking Deepfake Detection}

Existing deepfake methods primarily focus on manipulating the speaker during interactions, such as SHG \cite{prajwal2020lip,sun2025vividtalk} and speech synthesis \cite{jia2018transfer}. Correspondingly, speaking deepfake detection methods mainly rely on either unimodal or multimodal strategies to capture subtle discrepancies between authentic and manipulated samples.

Early unimodal detectors primarily focused on the visual modality \cite{li2020sharp,lai2025gm,ge2022deepfake,anand2025detecting}, identifying forged content by detecting inconsistencies in facial expressions \cite{mazaheri2022detection}, unnatural eye movements \cite{li2021exposing}, or misaligned lip synchronization \cite{yang2020preventing}. Subsequently, detection methods targeting manipulated audio began to emerge \cite{yi2023audio,li2024safeear,gu2025allm4add}. The ASVspoof challenges \cite{todisco2019asvspoof,delgado2021asvspoof} and the ADD challenges \cite{ADD2022,ADD2023} played an important role in advancing audio deepfake detection. 
Recent studies have emphasized leveraging the consistency between the audio and visual modalities to identify potential discrepancies\cite{wang2025audio,liu,hu2024,nie2024frade}. By examining the mismatch between facial movements and speech, researchers have proposed several methods \cite{Zhou_2021_ICCV,liu2026forgefinder,hu2024,zhang2024joint,wang2024building} that utilize joint feature embeddings and cross-modal attention mechanisms to enhance detection performance.

Although these methods achieve strong performance on existing datasets, they focus exclusively on speaker-oriented manipulations and overlook the listener, who plays an equally important role in interactive scenarios.

\subsection{Listening Head Generation}

Effective communication is a bidirectional process in which participants take turns assuming the roles of speaker and listener to exchange information. The speaker directly transmits information to the listener through verbal expression, while the listener actively considers the information provided by the speaker, decodes it, and conveys real-time feedback primarily through non-verbal behaviors. Therefore, 
LHG has attracted considerable attention
from researchers. Unlike SHG,
it aims to generate a listening head video
based on speaker’s audio and video as well as a reference listener
image. 

In early works, some data-driven
approaches \cite{feng2017learn2smile,nojavanasghari2018interactive}  based on facial keypoints were used to generate
2D listener motions, but they lost many details of facial expressions.
In recent years, many 3D-based methods have been developed
due to their excellent facial reconstruction capabilities. 
Zhou et
al. \cite{zhou2022responsive}  established a high-quality speaker-listener dataset, named
ViCo. 
At almost the same time, Ng et al. \cite{l2l} proposed a novel motion-encoding
VQ-VAE to learn a discrete latent representation
of realistic listener motion. 
Recently, several methods employ more advanced generative models such as diffusion models to further improve synthesis quality. 

Although LHG has enabled the development of many applications, including digital avatar generation and virtual anchors, it also makes the creation of listening deepfakes significantly easier, leading to potential negative impacts.

\begin{figure}[t]
	\centering
	\subfloat[]{
		\includegraphics[width=0.41\linewidth]{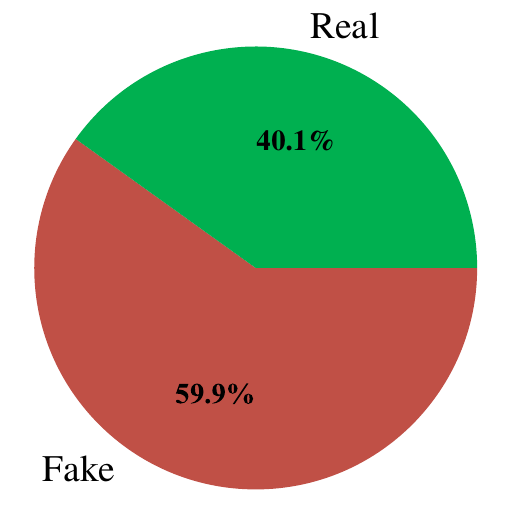}
		\label{f3_a}
	}	
	\subfloat[]{
		\includegraphics[width=0.52\linewidth]{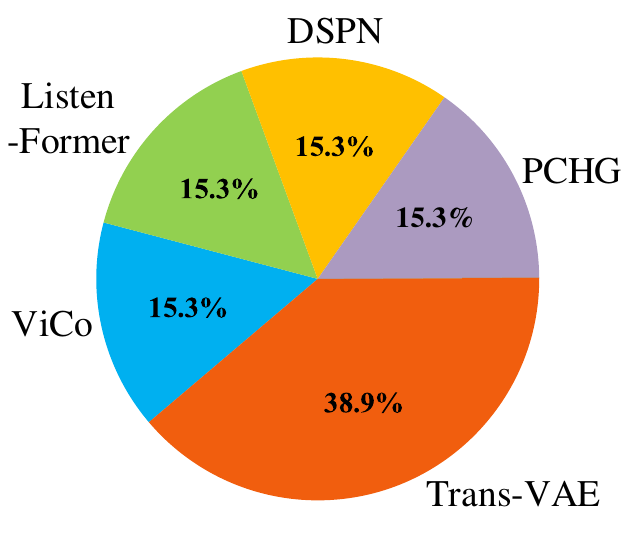}	
	}
	\vspace{-0.2cm}
	\caption{Distribution of the ListenForge dataset. (a) The ratio of real and fake samples. (b) The ratio of
		different methods used in forgery listening videos.}
	\label{f3}
	\vspace{-0.6cm}
\end{figure}

\section{ListenForge Dataset}

To the best of our knowledge, all existing public forgery datasets focus on manipulations of the speaker, and none are specifically designed for detecting listening deepfakes. To bridge this gap, we construct ListenForge, a dedicated dataset for listener-targeted forgery detection. The overall workflow is illustrated in Figure \ref{f2}.

\textbf{Data Collection.}
The dataset is constructed based on the ViCo \cite{zhou2022responsive} and NoXi \cite{cafaro2017noxi} corpora. These two audio-visual datasets are widely used for LHG tasks and contain videos captured from both the speaker’s and the listener’s viewpoints. ViCo consists of 893 online videos collected from YouTube. NoXi was recorded in a controlled laboratory environment, and we use 1,614 videos from its test set. Our dataset spans a diverse range of scenarios, incorporating not only well-known laboratory-recorded data but also real-world content sourced from YouTube.

\begin{figure*}[t]
	\centering
	\includegraphics[width=13cm]{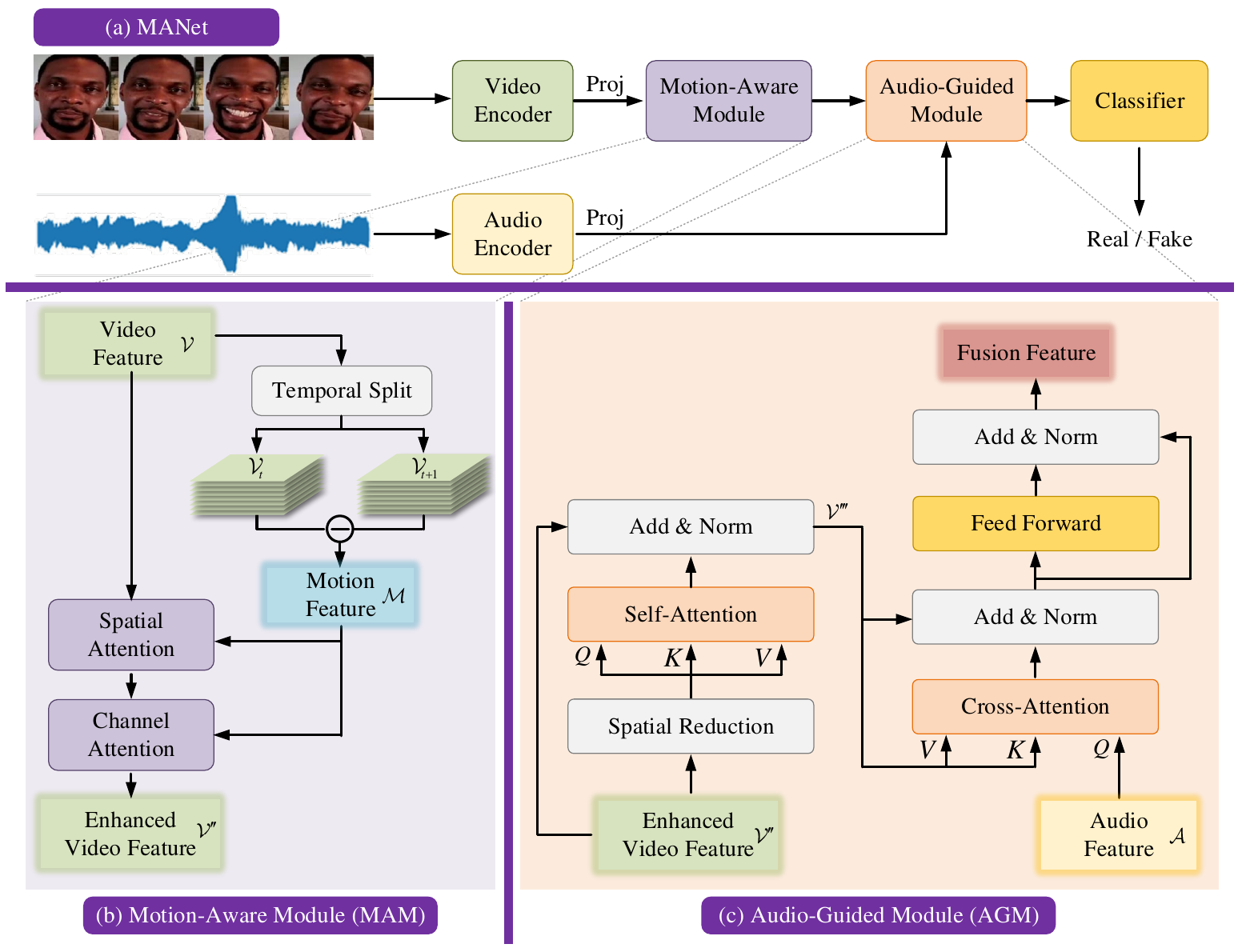}
	\caption{Illustration of the proposed MANet. (a) illustrates the overall framework. (b) illustrates the motion-aware module. (c) illustrates the audio-guided module.}
	\label{f4}
\end{figure*}

\textbf{Data Processing.}
We employ five listener head generation methods including ViCo \cite{zhou2022responsive}, DSPN \cite{dspn}, PCHG \cite{pchg}, Listenformer \cite{listenformer}, and Trans-VAE \cite{song2023react2023} to simulate realistic listener head movements. These methods represent all publicly available open-source approaches for LHG. ViCo, DSPN, PCHG, and Listenformer are applied to the ViCo dataset, while Trans-VAE is used with the NoXi dataset. 
Long video samples are segmented into 5-second clips, with any remaining tail shorter than 5 seconds merged into the preceding segment. Furthermore, to capture and analyze the listener's interactive responses to the speaker's content, this work appends authentic speaker audio to the listener video sequences, thereby constructing an audio-visual multimodal dataset tailored for listening forgeries.

\textbf{Data Statistics.}
Following the aforementioned processing and generation procedures, the finalized dataset comprises a total of 10,655 audiovisual clips. The dataset is partitioned into training, validation, and test sets, containing 8,746, 954, and 955 clips, respectively. The overall dataset distribution is illustrated in Figure~\ref{f3}.

\section{Method}

\subsection{Overview}

Given a video shot $x = ({x^v},{x^a})$ with the label $y$ where $x^v$ and $x^a$  denote the listener frame sequence, and speaker audio sequence respectively.  $y \in \{ 0,1\} $ represents whether $x$ is a forged listener video or not.  We formulate the listening deepfake detection problem as:

\begin{equation}
	\hat y = \mathcal{F}({x^v},{x^a})
\end{equation}
where $\mathcal{F}( \cdot )$ denotes the detection model with learnable parameters, and $\hat {y}$ denotes the predicted forgery probability.

We illustrate the proposed MANet framework in Figure \ref{f4}a. It consists of two pre-trained encoders that extract features from video frames and audio sequences. A motion-aware module and an audio-guided module are then introduced to fuse the features from both modalities, after which the fused representation is fed into a classification head for deepfake detection. 
The overall model is trained using cross-entropy loss:
\begin{equation}
	L = CE(\hat {y},y)
\end{equation}

\subsection{Motion-Aware Module}

During conversational interactions, a listener's feedback is mainly reflected through facial expressions and head poses, such as smiling, frowning, nodding, and head shaking. In listener forgery methods, the high-fidelity synthesis of these subtle motions remains highly challenging and frequently leaves behind perceptible artifacts. To more effectively capture motion-related forgery details within listener videos, we introduces a Motion-Aware Module (MAM) designed to focus on salient motion features. The architecture of MAM is shown in Figure \ref{f4}b.

\subsubsection{Module Structure}


In listener videos, subjects remain predominantly static. In contrast to the continuous lip movements observed in speaker videos, listener motions (e.g., blinking and nodding) are infrequent, typically swift, and subtle. Nevertheless, these subtle dynamics constitute the critical regions most susceptible to exposing artifact traces in forged listener videos. Therefore, the precise capture of dynamic forgery details within listener videos is crucial for enhancing detection accuracy. Motivated by this, our work leverages the temporal differences between adjacent frame-level visual features and applies attention mechanisms across both channel and spatial dimensions. 
As illustrated in Figure \ref{f4}b, let $\mathcal{V}\in\mathbb{R}^{N\times T\times C\times H\times W}$ be the visual features extracted by the visual encoder, where $N$ denotes the batch size, $T$ and $C$ represent the temporal dimension and the feature channels, and $H$ and $W$ correspond to the spatial dimensions. Given these features, we first employ a $1\times1$ convolution to compress the channel dimension by a reduction ratio $r$, thereby mitigating computational overhead. Taking $\mathcal{V}_t$ and $\mathcal{V}_{t+1}$ as an example, we compute their difference to obtain an approximate motion representation, denoted as $\mathcal{M}_t$. This differentiation process can be formulated as:

\begin{equation}
	{\mathcal{M}_t} = {\mathcal{V}_{t + 1}} - {\mathcal{V}_t}
\end{equation}


Subsequently, a joint spatial and channel attention module is introduced to amplify the dynamic forgery cues embedded within the original visual features. As illustrated in Figure \ref{f4}b, spatial attention (SPA) fusion is executed first, followed by the application of channel-wise attention (CHA). For brevity, the sequential cascade of this module is denoted as SCA. Formally, the MAM takes the motion-rich temporal difference features $\mathcal{M}$ as input to infer a 2D spatial attention weight $\mathcal{W}_s$ and a 1D channel attention weight  $\mathcal{W}_c$. These weights are then sequentially applied to the original visual features. The overall attention process can be formulated as follows:
\begin{equation}
	\mathcal{V'} = {\mathcal{W}_s}(\mathcal{M}) \otimes \mathcal{V}
\end{equation}

\begin{equation}
	\mathcal{V''} = {\mathcal{W}_c}(\mathcal{M}) \otimes \mathcal{V'}
\end{equation}
where $\otimes$ is element-wise multiplication, $\mathcal{V'}$ denotes the intermediate feature modulated by spatial attention and $\mathcal{V''}$ denotes the final output feature following spatial and channel enhancement. By reweighting the original features along both two dimensions, the model can amplify subtle forgery cues that might otherwise be overshadowed during standard forward propagation, thereby significantly enhancing its sensitivity to listener forgeries. In the following section, we will introduce the spatial and channel attention mechanisms, alongside their combined variants.

\subsubsection{Spatial Attention}
\begin{figure}[t]
	\centering
	\includegraphics[width=8.5cm]{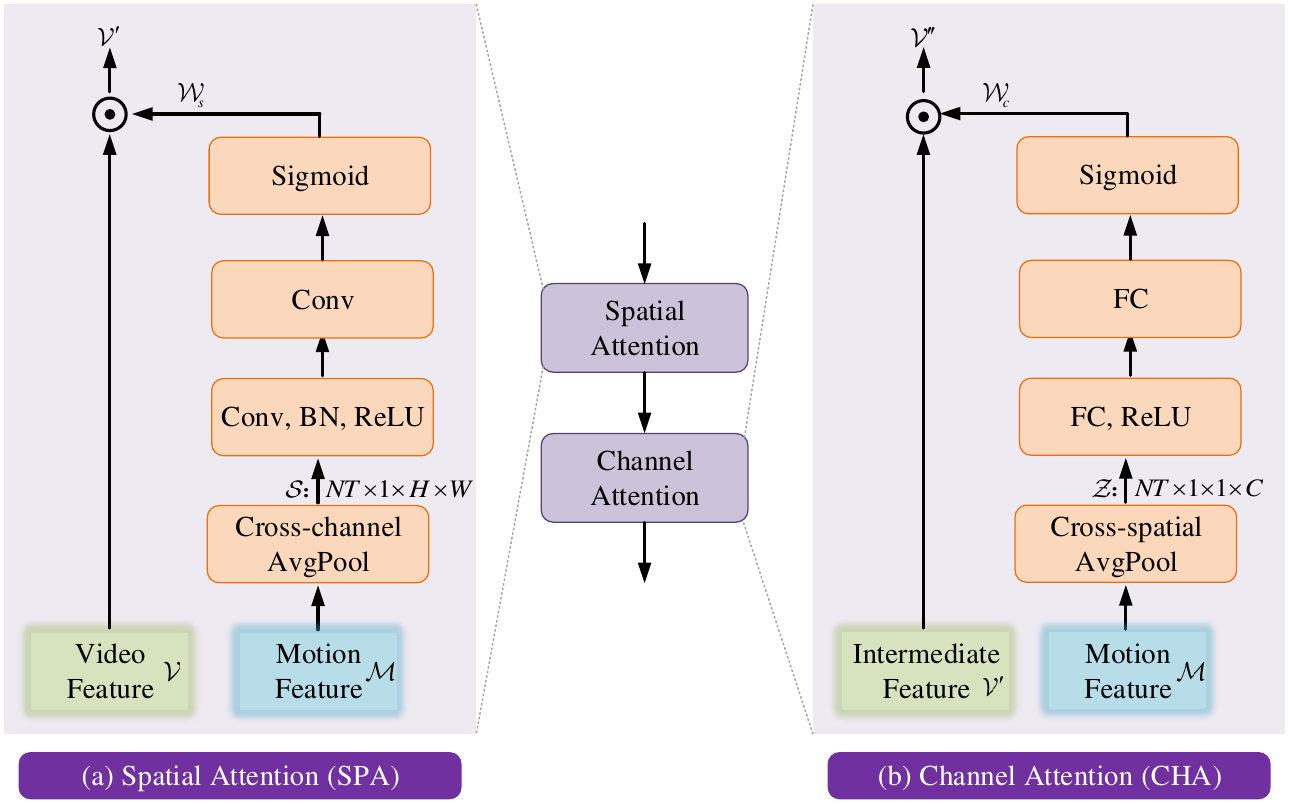}
	\caption{Illustration of the spatial and channel attention. }
	\label{sca}		
\end{figure}

As illustrated in Figure \ref{sca}a, 
we feed the video and motion features into the spatial attention module.  
In listener forgery scenarios, tampering anomalies are frequently concealed within the subtle dynamic variations of local facial regions. Consequently, relying solely on global features renders critical anomalous cues highly susceptible to being diluted or overshadowed. To address this, the spatial attention module employs explicit spatial modeling to guide the network in adaptively focusing on critical forged regions, thereby enhancing its spatial perception of fine-grained motion cues. Given the temporal difference features $\mathcal{M}$, we first perform average pooling along the channel dimension to aggregate the feature information distributed across all channels into a spatial descriptor $\mathcal{S}$. This operation can be formally expressed as:
\begin{equation}
	\mathcal{S}=\frac{1}{C}\sum_{i=1}^C\mathcal{M}_{1:W,1:H,i}
\end{equation}

Subsequently, a sequence of 2D convolutional layers processes the aforementioned spatial descriptor to comprehensively capture local spatial dependencies and further refine the attention distribution. Following mapping via a Sigmoid nonlinear activation function, the final spatial attention weight matrix, denoted as $\mathcal{W}_s$, is generated. Finally, this weight matrix is element-wise multiplied with the input video features (yielding the modulated features $\mathcal{V'}$). This operation selectively focuses on and amplifies critical regions exhibiting  appearance artifacts, such as incoherent or unnatural dynamic facial expressions.


\subsubsection{Channel Attention}

Following the spatial attention, this subsection introduces the channel attention mechanism. Since different feature channels generated by convolutional layers encode distinct semantic responses of the input data \cite{lei2019channel,chen2017sca}, introducing a gating mechanism to re-weight these channels can effectively amplify the representations highly correlated with forgery artifacts. As illustrated in Figure \ref{sca}b, The model first applies global average pooling  across the spatial dimensions of the temporal difference features $\mathcal{M}$ to aggregate a channel descriptor $\mathcal{Z}$ that encapsulates global motion dynamics. This process can be formulated as:

\begin{equation}
	\mathcal{Z} = \frac{1}{{W \times H}}\sum\limits_{i = 1}^W {\sum\limits_{j = 1}^H {\mathcal{M}_{i,j,1:C}} }
\end{equation}


Subsequently, to comprehensively capture the non-linear dependencies across channels, the module adopts a two-layer Fully Connected (FC) network including dimensionality reduction and dimensionality expansion operations, combined with a Sigmoid activation function, to improve its representation ability while limiting model complexity. It can be formulated as:

\begin{equation}
	{\mathcal{W}_c} = \sigma \left( {{W_1}\left( {\delta \left( {{W_0}(\mathcal{Z}) + {b_0}} \right)} \right) + {b_1}} \right)
\end{equation}
where ${W_0} \in {R^{C/r \times C}}$, $b_0 \in {R^{C/r}}$ and ${W_1} \in {R^{C \times C/r}}$, $b_1 \in {R^C}$ are learnable parameters. 
Through this mechanism, the network allocates higher activation weights to feature channels strongly associated with anomalous motion patterns, while simultaneously suppressing redundant channels irrelevant to deepfake detection. This adaptive channel recalibration process enables the model to focus on highly discriminative high-level semantic features, thereby further enhancing  the effectiveness of the feature representations.
\subsubsection{Variants}

To investigate the effects of different arrangement strategies for spatial and channel attention, this section introduces two variants of the SCA module: CSA and C//S. CSA reverses the computational sequence of the original module, applying channel attention followed by spatial attention. C//S adopts a parallel arrangement, specifically implemented by feeding the sum of the spatial and channel attention masks into a Sigmoid function, thereby constraining the activation weights to the interval of $(0,1)$. A detailed performance comparison of these structural variants will be discussed in Section 5.

\subsection{Audio-Guided Module}


In listening head generation methods, incorporating the speaker's contextual information is indispensable. Typically, both the audio and visual modalities of the speaker are provided as inputs. However, prior research \cite{listenformer} indicates that during the listening process, the speaker's audio stream conveys richer and more essential semantic cues compared to their facial movements. Consequently, in the ListenForge dataset constructed in this paper, the speaker's audio stream is integrated with the forged listener videos. Furthermore, this section introduces an Audio-Guided Module (AGM) designed to deeply explore the cross-modal interactive consistency between the speaker's audio and the listener's visual feedback, thereby effectively enhancing the model's performance in the LDD task.


The overall architecture of AGM is illustrated in Figure \ref{f4}c. In contrast to the symmetric fusion strategies commonly employed in speaker forgery detection—which treat audio and visual modalities equally—the proposed AGM adopts an asymmetric fusion paradigm. Specifically, it utilizes the listener's video as the primary modality and the speaker's audio as the guiding query. The core motivation behind this design is that, in LDD, the primary role of the speaker's audio signal is to provide a global "semantic context" rather than to establish strict spatio-temporal synchronization with the listener's visual modality. Consequently, the model leverages speaker's audio features as queries to adaptively retrieve and aggregate listener's  visual responses that are highly correlated with the current speech semantics. Specifically, prior to executing cross-modal fusion, the visual features $\mathcal{V}''$ are fed into a multi-head self-attention module to comprehensively model their temporal dependencies, yielding the refined visual features $\mathcal{V}'''$. Subsequently, a cross-attention mechanism is employed to facilitate interaction with the audio features $\mathcal{A}$ extracted by the audio encoder. In this operation, the queries are derived from $\mathcal{A}$, while the keys and values are supplied by $\mathcal{V}'''$. This computation can be formulated as follows:

\begin{equation}
	{\rm{Interact}}(\mathcal{V}''',\mathcal{A}) = {\rm{Softmax}}(\frac{{\mathcal{A}{W^Q} \cdot {{(\mathcal{V}'''{W^K})}^T}}}{{\sqrt d }})\mathcal{V}'''{W^V}
\end{equation}
where $W^Q$, $W^K$, $W^V$ are learnable parameters and $d$ is a scaling factor.

We expect the speaker’s audio feature to serve as queries in the cross-attention mechanism, reinforcing listener's visual features that are closely related to semantic content and thereby yielding a more comprehensive fused representation. Finally, a feed-forward layer is applied to produce the fused output. Residual connections and layer normalization are employed after both the attention and feed-forward layers to ensure training stability during the fusion process.

\section{Experiment Setup}

\subsection{Datasets and Evaluation Metric}
We evaluate our method on two benchmark datasets, including ListenForge, which we propose for listener-targeted facial forgery, and FaceForensics++ \cite{rossler2019faceforensics++}, a dataset designed for speaker-targeted facial forgery scenarios. We report the Area Under the Curve (AUC) and Accuracy (ACC) as evaluation metrics to assess our model on these datasets. 

\subsection{Implementation Details}

For visual data, we use a ResNet network pretrained on the ImageNet1K\_V1 \cite{imagenet15russakovsky}  dataset to extract visual features. For audio data, we adopt Wav2vec 2.0 \cite{baevski2020wav2vec} pretrained on LibriSpeech. The extracted visual and audio feature dimensions are 512 and 768, respectively, and the two modalities are temporally aligned. The proposed model is implemented in PyTorch and optimized using Adam on two RTX 3090 GPUs. The batch size for training all models is set to 8, with a learning rate of $1 \times 10^{-4}$ and a maximum of 20 epochs. The input audio is sampled at 16 kHz, and all input videos are resized to $224 \times 224$ pixels while retaining the RGB color channels for each frame.

\section{Results and Analysis}

\subsection{LDD vs SDD}

\begin{table}[t]
	\centering
	\caption{The AUC(\%) and ACC(\%) on of the baseline on the speaker-targeted and listener-targeted deepfake testsets.}	
	\renewcommand\arraystretch{1.2}
	\begin{tabular}{c|cc|cc}
		\hline
		\multirow{2}{*}{Method} & \multicolumn{2}{c|}{SDD} & \multicolumn{2}{c}{LDD} \\ \cline{2-5} 
		& AUC            & ACC            & AUC            & ACC            \\ \hline
		Baseline                & 69.09          & 77.57          & 92.03          & 81.88          \\ \hline
	\end{tabular}
	\label{t5}
	\vspace{-0.5cm}
\end{table}

This subsection presents a comparative analysis of the baseline model's performance when trained on listener forgery and speaker forgery datasets. Here, the "Baseline" refers to an architecture where a classification head is appended directly after visual feature extraction. For the speaker forgery scenario, we select FaceForensics++ \cite{rossler2019faceforensics++} as the dataset. This dataset is widely recognized as a standard benchmark within the deepfake detection domain and encompasses four facial manipulation techniques, rendering it comparable to ListenForge in terms of forgery diversity. As demonstrated in Table \ref{t5}, the baseline model achieves significantly superior performance on the ListenForge dataset compared to FaceForensics++. This discrepancy arises from the differing maturity levels of the two forgery techniques.
The speaker forgery methods employed in FaceForensics++ have undergone years of iteration, yielding highly photorealistic content with increasingly imperceptible forgery artifacts. Conversely, listening forgery technologies remain in their infancy, exhibiting imperfections regarding expression dynamics, interactive consistency, and the plausibility of contextual responses. This immaturity facilitates the model's ability to capture temporal or cross-modal anomalous cues, allowing the model to converge more readily and learn stable, discriminative features on the ListenForge dataset.
Furthermore, it is worth noting that FaceForensics++ was introduced in 2019; thus, the speaker forgery methods employed during its construction are now considered relatively outdated. Recently released datasets \cite{cai2024av,xu2024identity} typically leverage more advanced generative models, yielding highly deceptive speaker forgeries. This implies that current speaker deepfake detection tasks have actually become significantly more challenging. 

The cross-task comparative results not only highlight the disparity in technological maturity between these two forgery domains but also demonstrate that LDD currently presents a more discernible discriminative space. Most importantly, it indirectly confirms that LDD offers a promising research perspective—one that is highly complementary to traditional speaker deepfake detection—for tackling forgery detection in conversational scenarios.

\subsection{Comparative Experiments}

\begin{table}[t]
	\caption{The AUC(\%) and ACC(\%) of the models on ListenForge without retraining.}
	\renewcommand\arraystretch{1.2}
	\begin{tabular}{c|c|cc|cc}
		\hline
		\multirow{2}{*}{Method} & \multirow{2}{*}{Modality} & \multicolumn{2}{c|}{Valset}                            & \multicolumn{2}{c}{Testset}                           \\ \cline{3-6} 
		&                           & AUC                       & ACC                        & AUC                       & ACC                       \\ \hline
		Xception                & Visual                    & 60.45                     & 58.58                      & 62.02                     & 57.23                     \\
		MesoNet                 & Visual                    & \multicolumn{1}{l}{44.68} & \multicolumn{1}{l|}{50.39} & \multicolumn{1}{l}{43.28} & \multicolumn{1}{l}{46.99} \\
		CViT                    & Visual                    & \multicolumn{1}{l}{61.92} & \multicolumn{1}{l|}{45.07} & \multicolumn{1}{l}{56.64} & \multicolumn{1}{l}{52.15} \\ \hline
		AVTFD                   & Audio-Visual              & 58.76                     & 55.24                      & 54.30                     & 44.19                     \\
		MRDF                    & Audio-Visual              & 55.81                     & 62.47                      & 45.98                     & 50.47                     \\
		AVAD                    & Audio-Visual              & 47.09                     & 37.53                      & 55.18                     & 41.78                     \\ \hline
		MANet (Ours)                    & Audio-Visual              & \textbf{99.87}            & \textbf{98.53}             & \textbf{97.24}            & \textbf{89.74}            \\ \hline
	\end{tabular}
	\label{t1}
\end{table}

As we are the first to introduce the LDD task, there are currently no existing methods specifically designed for this task. To quantitatively evaluate the performance of our approach, we compare the proposed MANet with several existing  SDD models. The compared methods can be broadly categorized into unimodal and multimodal approaches. The unimodal methods include Xception \cite{rossler2019faceforensics++}, MesoNet \cite{afchar2018mesonet}, and CViT \cite{wodajo2021deepfake}, while the audio–visual methods include AVTFD \cite{liu}, MRDF \cite{zou2024cross}, and AVAD \cite{feng2023self}. 
Table \ref{t1} presents the evaluation results of the aforementioned models on the ListenForge dataset. The experimental results indicate that existing SDD methods exhibit inferior performance in both AUC and ACC metrics on ListenForge, falling short of the detection requirements for practical applications. In contrast, the proposed MANet achieves substantial improvements across all evaluation metrics on both the validation and test sets. Notably, on the test set, the AUC of MANet reaches 97.24\%, yielding remarkable performance gains of 42.94\%, 51.26\%, and 42.06\% over AVTFD, MRDF, and AVAD, respectively. These results not only expose the inherent limitations and inadequacy of traditional SDD models when tackling listener forgery tasks but also firmly validate the effectiveness of MANet's tailored architectural design for this specific challenge.


To further demonstrate the superiority of the proposed method, we additionally report the test results of existing models retrained on the ListenForge dataset. As shown in Table \ref{t2}, following retraining on the ListenForge dataset, the performance of all comparison models exhibits substantial improvement. For instance, the test AUC of MesoNet improves significantly from 43.28\% to 89.31\%. This phenomenon indicates that models originally designed for SDD can only become effective in this new domain after incorporating listener forgery data. This indirectly corroborates the necessity of constructing the dedicated ListenForge dataset.
Nevertheless, even under this more rigorous comparison setting, the proposed MANet consistently outperforms all retrained models across both AUC and ACC metrics. This validates the performance gains yielded by the core architectural innovations of our method. Notably, multimodal detection approaches generally surpass unimodal ones (e.g., on the test set, the detection AUC of AVTFD is 4.95\% and 3.03\% higher than that of Xception and CViT, respectively). This suggests that relying exclusively on visual information is insufficient for effectively capturing the discriminative features of listener forgeries; integrating audio semantics for contextual verification is highly beneficial for enhancing detection performance.
Among the multimodal detection methods, our proposed network continues to outperform AVTFD and MRDF. This further underscores the significance of the designed MANet architectural framework.



\begin{table}[t]
	\caption{The AUC(\%) and ACC(\%) of the models on ListenForge with retraining.}
	\renewcommand\arraystretch{1.2}
	\begin{tabular}{c|c|cc|cc}
		\hline
		\multirow{2}{*}{Method} & \multirow{2}{*}{Modality} & \multicolumn{2}{c|}{Valset}     & \multicolumn{2}{c}{Testset}     \\ \cline{3-6} 
		&                           & AUC            & ACC            & AUC            & ACC            \\ \hline
		Xception                & Visual                    & 97.92          & 98.05          & 88.24          & 87.23          \\
		MesoNet                 & Visual                    & 99.86          & 96.99          & 89.31          & 88.80          \\
		CViT                    & Visual                    & 97.97          & 97.59          & 90.16          & 87.96          \\ \hline
		AVTFD                   & Audio-Visual              & 99.39          & 98.43          & 93.19          & 89.28          \\
		MRDF                    & Audio-Visual              & 91.51          & 88.58          & 90.32          & \textbf{90.58} \\
		AVAD                    & Audio-Visual              & -              & -              & -              & -              \\ \hline
		MANet (Ours)                    & Audio-Visual              & \textbf{99.87} & \textbf{98.53} & \textbf{97.24} & 89.74          \\ \hline
	\end{tabular}
	\label{t2}
\end{table}

\subsection{Ablation Studies}

\subsubsection{Effectiveness of Motion-Aware Module}

In this section, we conduct ablation studies to validate the effectiveness of the motion-aware module within our framework. The results are summarized in Table \ref{t3}.
It indicates that, regardless of the fusion strategy employed, incorporating the motion-visual feature fusion module yields consistent performance improvements over the Baseline across most metrics on both the validation and test sets. This demonstrates that temporal difference features assist the model in more effectively capturing motion artifacts associated with listener reactions.
When comparing spatial attention (SPA) and channel attention (CHA), it is observed that CHA performs better on the validation set, whereas SPA exhibits superiority on the test set. This implies that spatial-level feature fusion is more advantageous for extracting highly generalizable features. Furthermore, coupling SPA and CHA in a sequential cascade (SCA) achieves superior detection performance compared to deploying either attention mechanism individually. This suggests that spatial and channel attention are functionally complementary; by jointly learning to enhance and suppress specific features, they effectively facilitate information propagation and fusion.
Regarding alternative arrangements of the attention modules, CSA yields the poorest performance, while SCA slightly outperforms the parallel C//S configuration. This discrepancy likely stems from the sequential order of dimensional focus: prioritizing the intrinsic spatial distribution of visual features before refining deep semantic representations along the channel dimension aligns more naturally with the inherent logic of feature extraction. 

\begin{table}[t]
	\centering
	\caption{The AUC(\%) and ACC(\%) on ListenForge for models using different motion-aware modules. Underlined values indicate sub-optimal results.}
	\vspace{-0.2cm}
	\renewcommand\arraystretch{1.2}
	\begin{tabular}{c|cc|cc}
		\hline
		\multirow{2}{*}{Method} & \multicolumn{2}{c|}{Valset}     & \multicolumn{2}{c}{Testset}     \\ \cline{2-5} 
		& AUC            & ACC            & AUC            & ACC            \\ \hline
		Baseline                & 98.71          & 93.51          & 92.03          & 81.88          \\
		+ CA                    & \textbf{99.32} & 91.82          & 95.36          & 79.58          \\
		+ SA                    & 98.56          & 91.20          & 95.36          & \textbf{84.40} \\
		+ SCA                   & {\ul 99.18}    & \textbf{94.34} & \textbf{95.43} & {\ul 84.08}    \\
		+ CSA                   & 98.01          & 91.20          & 94.42          & 78.53          \\
		+ C//S                  & 98.44          & 93.50          & 92.99          & 82.20          \\ \hline
	\end{tabular}
	\label{t3}
	\vspace{-0.2cm}
	
\end{table}

\subsubsection{Effectiveness of Audio-Guided Module}

\begin{table}[t]
	\centering
	\caption{The AUC(\%) and ACC(\%) on ListenForge for models using different audio-guided module.}
	\vspace{-0.2cm}
	\renewcommand\arraystretch{1.2}
	\begin{tabular}{c|cc|cc}
		\hline
		\multirow{2}{*}{Method} & \multicolumn{2}{c|}{Valset}     & \multicolumn{2}{c}{Testset}     \\ \cline{2-5} 
		& AUC            & ACC            & AUC            & ACC            \\ \hline
		Baseline + MAM          & 99.17          & 94.34          & 95.43          & 84.08          \\
		+ Spk\_Aud              & 99.05          & 94.03          & 96.51          & 82.93          \\
		+ Spk\_Aud + AGM        & \textbf{99.87} & \textbf{98.53} & \textbf{97.24} & \textbf{89.74} \\
		+ Spk\_Vid + AGM        & 99.79          & 97.48          & 95.88          & 85.55          \\
		+ Spk\_AV + AGM         & 99.88          & 98.43          & 95.80          & 83.77          \\ \hline
	\end{tabular}
	\label{t4}
\end{table}


This subsection investigates the impact of various fusion strategies for incorporating speaker information on the performance of MANet. As demonstrated in the second row of Table \ref{t4}, directly concatenating the speaker's audio features with the listener's visual features results in performance degradation across multiple evaluation metrics compared to the Baseline, yielding only a marginal improvement of 1.1\% in the testing AUC. Notably, such a straightforward concatenation strategy typically provides performance gains in SDD tasks \cite{Zhou_2021_ICCV}. This discrepancy suggests that treating speaker and listener features with equal weight during fusion in LDD tasks fails to effectively capture the semantic correlation between the speaker's audio and the listener's reactions. Consequently, it may introduce redundant or disruptive information irrelevant to the listener's visual content, thereby impairing the model's discriminative capability.
In contrast, when employing the proposed Audio-Guided Module (AGM) to fuse the speaker's audio features with the listener's visual features, the model achieves substantial improvements across all metrics. Specifically, on the test set, the AUC and ACC increase by 1.81\% and 5.66\%, respectively.
These results demonstrate that the effectiveness of incorporating speaker information for listener forgery detection is highly contingent upon the design of the cross-modal fusion mechanism. By designating the listener's visual modality as the primary stream and the speaker's audio as the auxiliary guide, our method successfully uncovers the underlying correlations between the two. This enables the model to more effectively capture the inconsistencies between the listener's behavior and the semantic context. which constitute the critical cues for detecting forged listener content.

Furthermore, we conduct a supplementary comparison evaluating the fusion of speaker and listener video features using the AGM module. As shown in the fourth row of Table \ref{t4}, utilizing the speaker's video for query guidance yields an improvement over the baseline on the test set (AUC 95.88\%, ACC 85.55\%), though it remains significantly inferior to the performance achieved using the speaker's audio. This demonstrates that while the speaker's visual signals offer some contextual semantic information, they are less informative than audio signals and introduce substantial irrelevant noise (e.g., background elements).
Moreover, providing both the speaker's audio and video as joint inputs did not lead to further performance gains. This limitation likely arises from a high degree of semantic redundancy between the video and audio modalities. The fused input not only increases the model's processing burden, but the spatial redundancy introduced by the video modality also tends to obscure the pure semantic cues provided by the audio modality.

\subsection{Qualitative Results}

\begin{figure}[t]
	\centering
	\includegraphics[width=8cm]{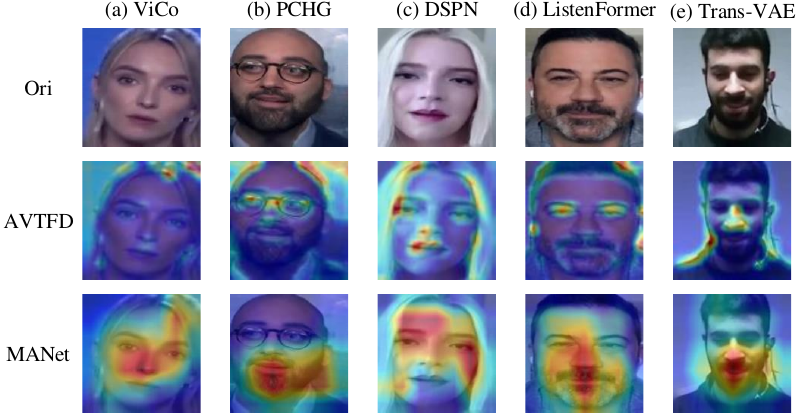}
	 \vspace{-0.3cm}
	\caption{Visualization of the attention maps. The redder the color, the higher the attention allocated to that region by the model during the inference process.}
	\label{am}	
	 \vspace{-0.7cm}	
\end{figure}

To provide a more intuitive validation of the proposed method's effectiveness, this subsection randomly selects samples generated by five different forgery methods from the ListenForge test set, and visually compares the visual attention maps of the proposed MANet with the existing speaker forgery detection method AVTFD. As shown in Figure \ref{am}a, AVTFD tends to focus primarily on the background or unmanipulated regions, while MANet accurately localizes the forged facial areas. In subfigure (b), MANet precisely targets the visibly unnatural chin region, demonstrating its high sensitivity to subtle anomalies in facial movement. In subfigure (c), MANet attends to the overall facial motion pattern while assigning a lower weight to the relatively static mouth region; conversely, in subfigures (d) and (e), the model highlights the unnatural details of the smile. These observations indicate that, MANet can dynamically adjust its focal points based on the semantic context. This enables a more targeted identification of listener forgeries, ultimately yielding superior detection performance on the listener forgery dataset.

\subsection{Conclusion}

This paper introduces listening deepfake detection, a new and underexplored problem that extends deepfake analysis beyond the conventional speaking-centric paradigm. We show that listening forgeries exhibit distinct characteristics and that existing speaker-targeted detection methods are ineffective in this scenario.
To support research in this direction, we construct ListenForge, the first dataset specifically designed for the LDD task. Furthermore, we propose MANet, which combines a motion-aware module to capture subtle temporal inconsistencies in listener behavior with an audio-guided module that leverages semantic correlations between speaker audio and listener responses.
Extensive experiments demonstrate that MANet consistently outperforms existing unimodal and multimodal methods, even after retraining. 
In future work, we plan to investigate unified detection frameworks that jointly model speaking and listening behaviors, enabling synchronized detection of speaking and listening forgeries within a single interactive system. 

\begin{acks}
	This work is supported in part by National Nature Science Foundation
	of China (No.62071039) and in part by Beijing Natural Science Foundation
	(No.L223033).
\end{acks}
\bibliographystyle{ACM-Reference-Format}
\bibliography{sample-base}

\end{document}